\documentclass{bmvc2k}

\usepackage{amsmath}
\title{Learning Gaussian Maps for Dense Object Detection}

\addauthor{Sonaal Kant}{sonaal@paralleldots.com}{1}

\addinstitution{
 Paralleldots Inc.\\
 https://www.paralleldots.com/ \\
}

\runninghead{SONAAL}{Learning Gaussian Maps for Dense Object Detection}

\begin{document}

\maketitle

\begin{abstract}
Object detection is a famous branch of research in computer vision, many state of the art object detection algorithms have been introduced in the recent past, but how good are  those object detectors when it comes to dense object detection? In this paper we review common and highly accurate object detection methods on the scenes where numerous similar looking objects are placed in close proximity with each other.  We also show that, multi-task learning of gaussian maps along with classification and bounding box regression gives us a significant boost in accuracy over the baseline. We introduce Gaussian Layer and Gaussian Decoder in the existing RetinaNet network for better accuracy in dense scenes, with the same computational cost as the RetinaNet. We show the gain of 6\% and 5\% in mAP with respect to baseline RetinaNet.
Our method also achieves the state of the art accuracy on the SKU110K \cite{sku110k} dataset.

\end{abstract}

\section{Introduction}
Computer vision as a field has grown from research to more of an applied field. Many industries are using computer vision either to enhance their existing technology or creating an altogether new product around it. Either way, object detection algorithms play a crucial role in almost every aspect. It has attracted much attention in the computer vision field because of its numerous real world applications, from Self driving cars to Surveillance many applications require an object detection algorithm. Similar to these, companies are also using object detection in retail stores, to maximize sales and store inventory management.  Recent work from Fuchs et al. \cite{holoselecta} shows the computer vision challenges in supermarket or retail stores environment.  They have also shown results from  transfer  learning  for  image-based product classification and multi-product object detection, using multiple  CNN  architectures on the images of vending machines.
 
Unlike popular object detection datasets such as ILSVRC \cite{deng2012ilsvrc}, PASCAL VOC \cite{hoiem2009pascal} detection challenges, MS COCO \cite{lin2014microsoft}, and the very recent Open Images v4 \cite{kuznetsova2018open} the retail stores based datasets such as  \cite{sku110k}  \cite{WebMarket} is more densely packed. The annotations for the WebMarket \cite{WebMarket}, CAPG-GP \cite{FGC-OSL} and Grocery Products \cite{GP} have been released by  \cite{varadarajan2019benchmark} for more robust comparison of object detection algorithms on a benchmark of retail stores datasets. The problem while working on the densely packed datasets is that the very similar looking objects are placed in close proximity with each other which makes it difficult for the object detection algorithm to find the boundaries and hence result in many overlapping bounding boxes with high objectness score. 

Object detection algorithms have evolved in many years, starting off with the Two stage detection method RCNN \cite{girshick2014rich} and its faster successors such as FastRCNN \cite{girshick2015fast} and FasterRCNN \cite{ren2015faster} which introduced a region proposal network (RPN). This later was improved by Mask-RCNN \cite{he2017mask} by adding a segmentation output as a multi-task learning approach. Evolving from Two stage to single stage detection for better and faster results, YOLO \cite{redmon2016you}, SSD \cite{ssd}, and YOLO9000 \cite{redmon2017yolo9000} were introduced which removed the need of proposals from the algorithms.   Recent work from  \cite{sku110k} shows how badly the standard object detection methods fail in the case of densely packed scenes. Lin et. al \cite{lin2017focal} shows with his research that the foreground-background class imbalance is the reason why these state of the art detectors perform poorly. In order to handle the class imbalance and scale variance he introduces feature pyramid network (FPN) \cite{lin2017feature} with focal loss. 

\textbf{Contributions}   We extend the work of Lin et al. \cite{lin2017focal} by adding an auxiliary loss to the existing RetinaNet architecture. We show that sharing representations between related tasks can enable our model to generalize better on our original task. We introduce a gaussian loss as an auxiliary branch for predicting a low resolution, per-pixel heat-map, describing the likelihood of a object occurring in each spatial location, in parallel with the existing branch for bounding box regression and object classification. We try to overcome the limitation of detecting objects in close proximity by 
enforcing the network to learn less likelihood for the pixels which are not the centers of object, hence making it easier for the anchors to learn the boundaries. We introduce two different network architectures to emphasize the importance of multi-task learning in object detection for densely packed scenes. 
\begin{enumerate}
    \item Gaussian Decoder Network is a multi-task learning approach where the backbone parameters are shared, with two different decoders. The second decoder is for gaussian map learning that can help the model focus its attention on those features that actually matter which will provide additional evidence for the relevance or irrelevance of those features for better anchors. 
    \item Gaussian Layer Network is an optimized version of Gaussian Decoder Network with less parameters. In this, we have introduced a gaussian layer on top of a shared decoder for gaussian maps prediction which helps the network to learn precise anchors. 
\end{enumerate}
Both architectures show the improvement in accuracy in SKU-110K \cite{sku110k} and other groceries dataset such as WebMarket \cite{WebMarket}, GroceryProducts \cite{GP}, CAPG-GP \cite{FGC-OSL} with the baseline.

\begin{figure}
  \includegraphics[width=\linewidth]{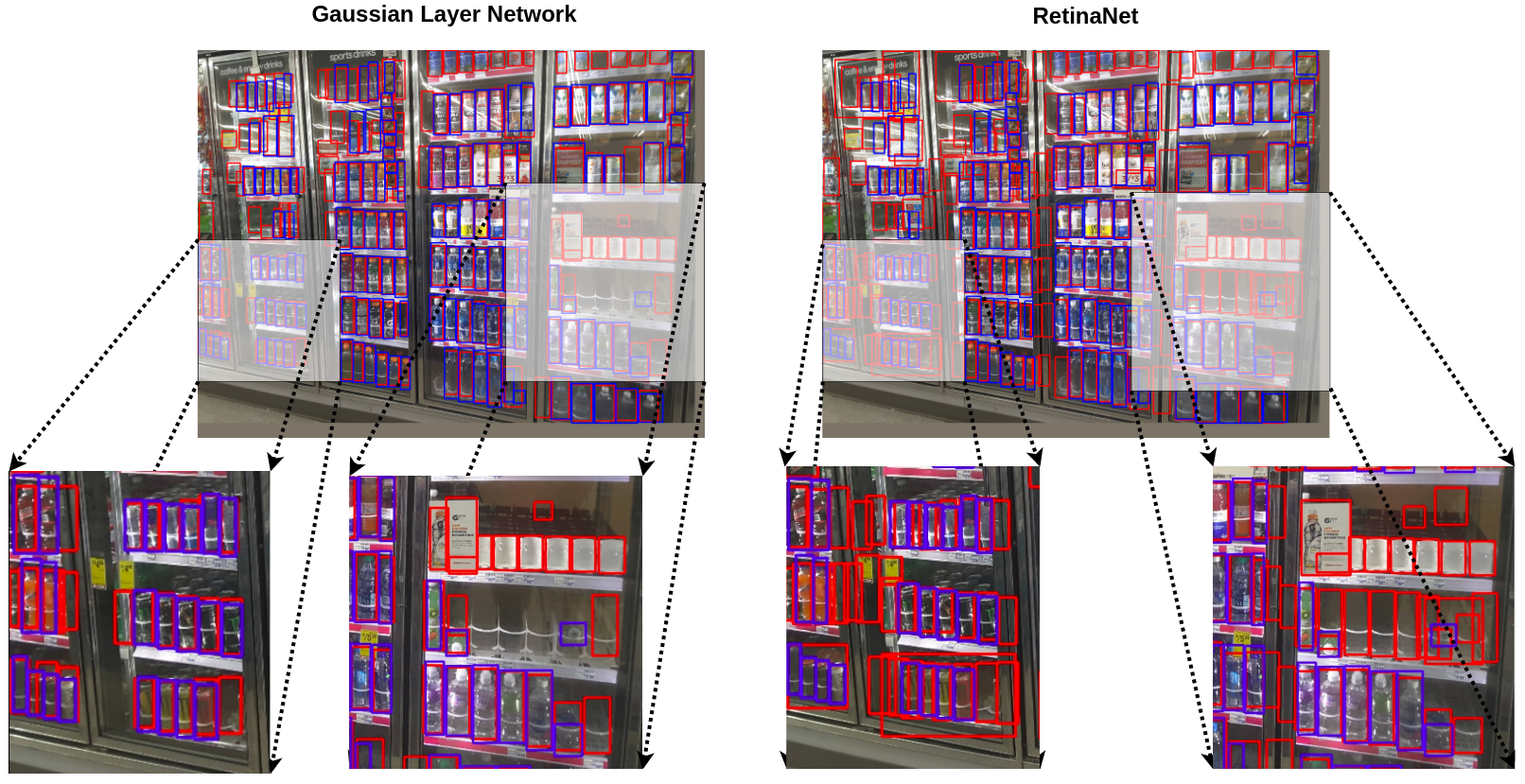}
  \caption{\textbf{Example Prediction In SKU-110K Dataset}. (Left) Detection results of Our Method. (Right) Detection Results of baseline RetinaNet. The red bounding box is the prediction and the blue bounding box is the groundtruth.}
  \label{fig:sample}
\end{figure}

\section{Related Work}
Goldman et. al \cite{sku110k} has recently released their SKU-110K dataset. The dataset represents various possible dense object detection examples at different scales, angles and containing different types of noise. Different brands and products which are often distinguishable only by fine-grained differences are kept in close proximity with each other. It contains 8233 training images, 588 validation images and 2941 testing images containing objects of different aspect ratios, quality and different lighting conditions. This variety in the dataset makes it a good benchmark to evaluate the performance of the object detection algorithm in densely packed scenes. 
They have also shown the performance of state of the art object detection algorithms such as Faster-RCNN \cite{ren2015faster}, YOLO9000 \cite{redmon2017yolo9000} on their dataset to compare it with their approach. They have extended their work on RetinaNet \cite{lin2017focal} by introducing another head along with a classifier and the bounding box regressor and they call it a Soft-IoU layer.
They propose that the classifier predicts the objectness score which is not sufficient in dense images because of the multiple overlapping bounding boxes which often reflect multiple tightly packed objects. So to handle these cluttered predictions they introduce IoU score as an additional value for every predicted bounding box with the object. In order to handle the multiple predicted bounding box they introduce an approach that replaces Non Maximum Suppression (NMS) with EM-Merger. EM-Merger takes the predictions as a set of 2D Gaussians and performs Gaussian Mixture Modeling, returning the final set of predictions.
In our approach we train the model end-to-end, to predict the set of 2D gaussians which will ultimately help to learn to give better anchors instead of relying on the naive post processing step.

\textbf{Instance Segmentation}. The research community has started shifting their attention to the more complex task of instance segmentation, while the object detection methods give the bounding box for each object, the segmentation models give the pixel-level mask for that object. K He et al. \cite{he2017mask} comes up with the multi-task approach of training an object detection method along with the instance segmentation. They called their multi-task architecture as MaskRCNN which has an additional branch for predicting segmentation masks on each Region of Interest (RoI) in a pixel-to pixel manner. This multi-task training approach proves to be better than the normal FasterRCNN \cite{ren2015faster}, hence giving the accuracy boost on both the object detection and Instance segmentation tasks. Path Aggregation Network extends the idea of MaskRCNN by introducing the bilinear Interpolation in their ROI Align module. Following the idea of training object detection and segmentation, Fu et al. \cite{fu2019retinamask} introduces RetinaMask which is the extended version of RetinaNet. In our approach where the objects are densely placed together, predicting the class at every pixel makes it difficult for the network to learn the objectness. So instead we only predict the gaussian with maxima at the centre of the object. 

\textbf{Gaussian Based detection}.  Many problems such as Human Pose Estimation  \cite{zhang2019distribution}  \cite{peng2018jointly}, Face Keypoint detection  \cite{moon2019posefix} etc. uses gaussian maps in their approach. Similar to ours, Baek et. al \cite{baek2019character} uses gaussian map prediction to localize the characters in scene text localization tasks. The performance of their algorithm shows the effectiveness of this method. They train a weakly supervised method to detect each individual character using a gaussian map which they call Region score. Despite being a completely different problem it relates to our retail based densely packed scenes. The numerous similar looking characters which are stacked together in a word correlates with the products which are placed together on the shelf. Inspired by his idea we add this gaussian map to our RetinaNet object detection. 

\section{Motivation}
Many Object Detection algorithms and their variants have been proposed for object detection tasks like PASCAL \cite{hoiem2009pascal}, COCO \cite{lin2014microsoft} but the object detection in dense scenes is still an area which is not much explored. Retail stores and supermarkets are the perfect case in point for  densely packed scenes. They contain similar looking products which are very large in number and placed in close proximity with each other. Recent study by  \cite{sku110k} has shown the state of the art object detectors like YOLO \cite{redmon2016you}, Faster-RCNN \cite{ren2015faster}, fails to perform well when it comes to dense object scenes. The performance was drastically improved by the RetinaNet \cite{lin2017focal} architecture with focal loss because of its ability to handle positive-negative class imbalance while training.  Goldman et al. \cite{sku110k} work on top of the RetinaNet architecture introduces an EM-Merger module, a gaussian mixture model to merge the predictions from retina net. Building upon their idea of treating the predicted boxes as a 2d gaussian, we hypothesize that instead of using a post processing method, adding an auxiliary loss of gaussian map to the RetinaNet architecture and performing a multi-task learning approach will directly help the anchors to learn the better boundaries of the object and will also help the network to generalize better. To validate our hypothesis we use RetinaNet \cite{lin2017focal} as our baseline object detector and we extend this by adding an auxiliary gaussian loss on the encoder and on both the encoder and decoder.

\begin{table*}
\centering
\begin{tabular}{lllllll}
\hline
Method     & AP    & AP$^{.50}$ & AP$^{.75}$ & AR$_{300}$ & AR$_{300}^{.50}$  \\ \hline
           
Faster-RCNN  \cite{sku110k} & 0.045 & - & 0.010 & 0.066 & - \\ 
                    YOLO9000  \cite{sku110k} & 0.094 & - & 0.073 & 0.111 & - \\ 
                    RetinaNet  \cite{sku110k} & 0.455 & - & 0.389 & 0.530 & - \\ 
                    Goldman et. al  \cite{sku110k} & 0.492 & - & 0.556 & 0.554 & - \\ 
                    Goldman et. al*  \cite{sku110k} & 0.514 & 0.853 & 0.569 & 0.571 & 0.872 \\ \hline
                    MaskRCNN & 0.403 & 0.742 & 0.396 & 0.465 & 0.778 \\                       
                    \emph{Gaussian Decoder}  & 0.512 & 0.878 & 0.552 & 0.582 & 0.917 \\
                    \emph{Gaussian Layer}  & \textbf{0.521} & \textbf{0.891} & \textbf{0.562} & \textbf{0.596} & \textbf{0.931} 
                       \\ \hline
\end{tabular}
\caption{Performance of $our$ approach on SKU-110K dataset. We compare our model also with the baselines provided by \cite{sku110k}. * denotes results obtained using the improved model given by the authors at \href{https://github.com/eg4000/SKU110K_CVPR19/issues/9}{URL}}
\label{table:sku110k}
\end{table*}

\section{Baseline}
We use RetinaNet \cite{lin2017focal} as our baseline as it has been proven to work better than Faster-RCNN. The reason for this is, Faster-RCNN uses Region Proposal Network for bounding box regression and classification on top of high level feature map which losses lots of semantic information thus unable to detect small objects, while RetinaNet uses Feature Pyramid Network (FPN)
that naturally leverages the pyramidal shape of a Convnet feature hierarchy while creating a feature pyramid that has strong semantics at all scales, hence solving the problem of detecting small objects. The class imbalance is another reason why we use RetinaNet as our baseline, many object detection algorithm faces the problem of huge class imbalance because of less positive anchors and very large number of negative anchors. Research work like OHEM \cite{OHEM} has also addressed this problem while training the object detection methods, to overcome this problem RetinaNet uses FocalLoss for classification. Focal loss is an extension of cross entropy loss that down-weights the loss assigned to easy negatives hence preventing the easy negatives to harm the detector during training.

 \[pt =
  \begin{cases}
    p       & \quad \text{if } y \text{ = 1,}\\
    1 - p  & \quad \text{otherwise}
  \end{cases}
  \]

\begin{equation}
  L_{cls}(pt) = \alpha (1 - pt)^\gamma \log(pt)
  \label{eq:cls}
\end{equation}


We use $\gamma$ and $\alpha$ as mentioned in the original paper that is 2, 0.25 respectively. 
Unlike classification subnet that uses focal loss instead of cross entropy loss, the bounding box subnet uses the standard Smooth L1 loss that is applied on all positive anchors. 

\begin{equation}
    L_{reg} =
  \begin{cases}
    0.5x^2       & \quad \text{i } |x| \text{ < 1,}\\
    |x| - 0.5  & \quad \text{otherwise}
  \end{cases}
  \label{eq:reg}
\end{equation}

\section{Our Approach}
Baek et al \cite{baek2019character}  uses gaussian heatmap for predicting the character level bounding boxes, for scene text detection. The scene text detection datasets have numerous number of words and in a word the characters are close together and almost similar looking. This trend in the scene text dataset can be seen in our densely packed scenes. Similar to  that, we can see our objects as a 2D gaussian with its peak at the center of the object and the $sigma_x$ and $sigma_y$   of that gaussian is defined by the width and the height of the object. Unlike scene text dataset where they don’t have the character level bounding box annotation which inspired him to do weakly supervised character detection, we have the bounding box annotation for every object so we perform a fully supervised training by generating the gaussian heatmaps using the ground truth bounding box. For each training image, we generate the ground truth gaussian map using the object ground truth bounding box. The gaussian map is a set of 2d gaussians for every object in the training image, every 2d gaussian represents the object with the highest probability at the center of the object. To generate the gaussian map, we first make a square gaussian of size 120 and sigma 40. For every bounding box in the training image we find the homography $\mathbf{H}$ using four point transform $\mathbf{P}$, which is then applied to the gaussian $\mathbf{G}$ to wrap it to the box area.
We consider $N$ ground truth bounding boxes $B_i \in R^2$ and convert them to $2D$ gaussians, we start with an empty image $\mathbf{I}$

  \[ \mathbf{G} = \exp^{-4\log2((x - x_c)^2 + (y - y_c)^2) / \sigma^2}\]
  \[ \mathbf{H} = \{h_i\}_{i=1}^{N} = \{P(B_i)\}_{i=1}^{N} \]
  \[ \mathbf{I} = \mathbf{I} + \mathbf{H}(\mathbf{G}) \]

Baek et. al.  \cite{baek2019character} uses VGG as his backbone and Unet based architecture for training. Following his work, we add the UNet architecture to our RetinaNet Baseline. 
We train RetinaNet from scratch with Resnet50  \cite{he2016deep} as the backbone which is pretrained on ImageNet \cite{deng2009imagenet}. We try two different approaches to test the hypothesis of adding Gaussian Maps. We would like to emphasize on the point that we do multi-task learning training with additional Mean Squared Loss(MSE) with hard example mining on the output gaussian map  $\textbf{I}^*$ added to the existing RetinaNet architecture and we call this a gaussian loss. We create two empty masks, $\delta_n$ for negative sampling and $\delta_p$ for positive sampling, of dimension \textbf{I}. $\delta_n$ is activated when the target is less than equal negative thresh and $\delta_p$ is activated when it is greater than equal positive thresh.

\begin{equation}
    \mathcal{L}_{gl} = \dfrac{1}{n} \sum_{i=1}^{n}  \sum_{xy} ( \delta_n || \textbf{I}_{ xy} - \textbf{I}_{xy}^* ||^{2} + \delta_p || \textbf{I}_{xy} - \textbf{I}_{xy}^* ||^{2} )
    \label{eq:gl}
\end{equation}

\begin{figure}
  \includegraphics[width=\linewidth]{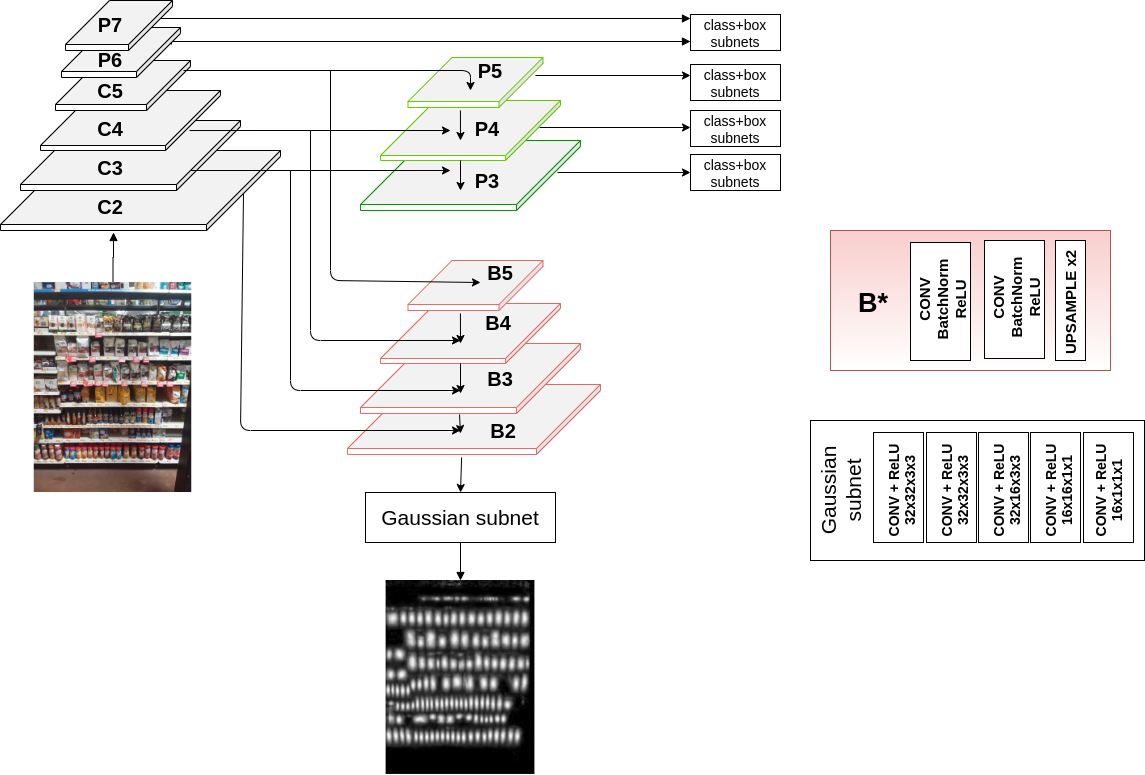};
  \caption{Gaussian Decoder Network. A standard UNet Achitectecture is used with B2, B3, B4, B5 layers as a decoder, the intrinsic details of each layer can be seen in B*.}
  \label{fig:gdn}
\end{figure}

\subsection{Gaussian Decoder Network (GDN)}

Gaussian Decoder Network is an extended version of RetinaNet. Similar to RetinaNet we use Resnet50 \cite{he2016deep} as the encoder but instead of using the feature pyramid network as the decoder we propose a separate decoder which predicts the sets of 2d gaussians of every objects in the image. The feature pyramid network predicts the bounding box and the classes at every level. It combines low resolution, semantically strong features with the high resolution semantically weak features using lateral skip connections. In order to predict the gaussian center of each product in the image, the network should have an idea of “what” and “where”, which means what are the objects which the network has to predict and where are the center of those objects, for this Ronneberger et. al. \cite{ronneberger2015u} showed that the U shape architecture which has a contracting path mainly consists of convolutional and downsampling layers and the expansive path which consist of transpose-2d convolutional layers for upsampling along with the skip connections that are used to concatenate the features from contracting path to the expansive path is a good architecture.
The same idea is used to design the decoder of the GDN. As shown in Figure \ref{fig:gdn}, layers C2, C3, C4, C5 of the encoder Resnet50  \cite{he2016deep} are used as skip connections to the decoder. The layers B2, B3, B4, B5 in decoder consists of convolution, batchnorm and relu followed by an interpolation of 2x.The interpolated output from B2 of size H/2, W/2 is then pass to gaussain subnet for gaussian map prediction.

\subsection{Gaussian Layer Network (GLN)}

We propose a Gaussian Layer in the RetinaNet architecture with lesser parameters and better accuracy as compared to the Gaussian Decoder. Gaussian Layer Network is a multitask learning architecture with shared encoder and decoder. The anchors from layers P3, P4, P5, P6, P7 are trained using the standard regression loss. An additional gaussian loss is applied on the output of the gaussian subnet. We hypothesize that simultaneously training the anchors with both gaussian and regression loss will lead to a more accurate bounding boxes. We take the concatenation of low level features C2 and P3 as the input for the gaussian layer ($B2$). The outputs of the gaussian layer are passed to the gaussian subnet.

\begin{figure}
  \includegraphics[width=\linewidth]{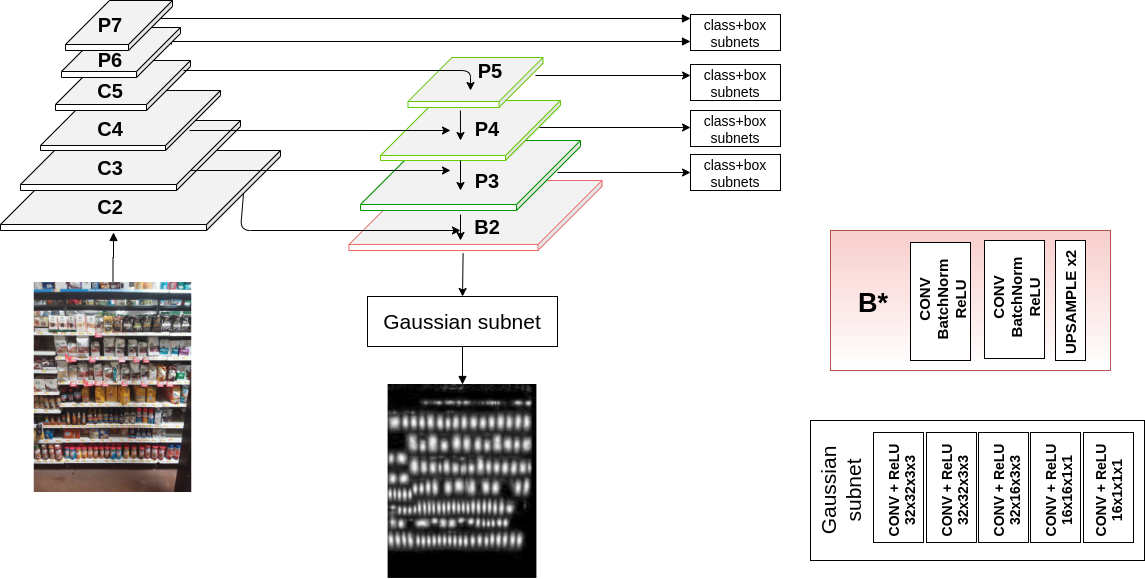}
  \caption{Gaussian Layer Network. Instead of adding an additional decoder for gaussian map we add an extra layer B2 and call it as a gaussian layer. The details of the gaussain layer can be seen in B*. }
  \label{fig:gln}
\end{figure}

Applying gaussian loss will not only refine the anchors from P3, P4, P5 but will also enhance the low level features C2. Similar to the bbox subnet and class subnet introduced in RetinaNet, we introduce a gaussian subnet which has the sequence of convolution, batchnorm and relu blocks as shown in Figure \ref{fig:gln}. The final output from the gaussian layer is a single channel map of size (H/2, W/2) where H and W is the height of the original image. The final loss is calculated as the weighted sum of classification (\ref{eq:cls}), regression (\ref{eq:reg}) and gaussian loss (\ref{eq:gl}).

\begin{equation}
    \mathcal{L}_{total} = \lambda_1\mathcal{L}_{cls} + \lambda_2\mathcal{L}_{reg} + \lambda_3\mathcal{L}_{gl}
\end{equation}

\section{Experiments \& Results}
We train one model of each proposed approach on the SKU-110K training set which has 8233 images and use the checkpoint with best performance on the validation set that has 556 image. These models trained on SKU-110K \cite{sku110k} dataset are tested not just on the test set of SKU-110K but also on WebMarket \cite{WebMarket}, GroceryProducts \cite{GP}, CAPG-GP \cite{FGC-OSL} and Holoselecta \cite{holoselecta}. All the implementation is done in Pytorch \cite{NEURIPS2019_9015}. We compare our method on SKU-110K datasets with the baselines and the full approach given by Goldman et. al \cite{sku110k} and we also add another baseline of MaskRCNN to compare our method with the multi-task segmentation approach. We also compare the full approach given by Goldman et. al \cite{sku110k} on other retail based datasets. We want to make clear that we used the improved weights given by the author which are better than the one he reported in the paper \cite{sku110k}. The link of those weights can be found \href{https://github.com/eg4000/SKU110K_CVPR19/issues/9}{here}.

\subsection{Training}
We train all our experiments by following the settings given in the original RetinaNet \cite{lin2017focal} paper. Input images are resized by keeping minimum dimension as 800 and maximum dimension as 1333. All our models are trained on a single 1080ti GPU, as some of the models take larger GPU RAM we keep the batch size as 1 for all training. To compare our models well with the previously trained methods we keep all the hyperparameters constant as mentioned in the original paper. We take the anchor boxes on feature pyramid levels P3 to P7. Every anchor box is matched with a single ground truth bounding box and all the anchors that have intersection over union overlap greater than 0.5 are taken as positive anchors and those with less than 0.4 are taken as negative, rest all the anchors are ignored from training. We then train our network with Focal Loss for classification, regression for bounding box and L2 Norm with hard example mining for gaussian maps till the best validation loss is not achieved. 
We have also shown MaskRCNN \cite{he2017mask} as our one of the baselines, we have used the implementation provided by Pytorch \cite{NEURIPS2019_9015} with Resnet 50 as the backbone which is common for all the networks we have trained as well the current state of the art model on SKU-110K \cite{sku110k} dataset.

\subsection{Comparison on SKU-110K}
SKU-110K test set comprises of 2941 images with total 432,312 ground truth bounding boxes which makes it approximately 146 objects per image, similar statistics belong to the training set. We compare our models with baselines provided by  \cite{sku110k}. We also add MaskRCNN as another baseline to the list for future work comparisons. As shown in Table \ref{table:sku110k}, Gaussian Decoder and Gaussian Layer Network outperforms the baseline RetinaNet with approximately 5\% and 6\% respectively. This accuracy gain on the baseline validates our hypothesis of performing multitask learning with gaussian maps.We also show the improvement in accuracy in comparison with the previous state of the art method. We want to make it clear that our final accuracy is better than the numbers reported in the paper by 3\% and also than the weights given by the author in his github repository by 0.8\%.   

\begin{table*}
\centering
\begin{tabular}{|l|l|l|l|l|l|l|}
\hline
Dataset                & Method     & AP    & AP$^{.50}$ & AP$^{.75}$ & AR$_{300}$ & AR$_{300}^{.50}$  \\ \hline
WebMarket              & Goldman et. al* & 0.383 & 0.773 & 0.332 & 0.491 & 0.855  \\ 
                       & \emph{Gaussian Decoder}  & 0.397 & 0.798 & 0.340 & 0.547 & 0.946 \\
                       & \emph{Gaussian Layer}  & \textbf{0.403} & \textbf{0.813} & \textbf{0.340} & \textbf{0.551} & \textbf{0.954}  
                       \\ \hline
Holoselecta            & Goldman et. al* & \textbf{0.454} & \textbf{0.835} & \textbf{0.447} & \textbf{0.581} & \textbf{0.955}  \\ 
                       & \emph{Gaussian Decoder}  & 0.368 & 0.717 & 0.316 & 0.497 & 0.842 \\
                       & \emph{Gaussian Layer}  & 0.384 & 0.705 & 0.368 & 0.524 & 0.843 
                       \\ \hline
GP                     & Goldman et. al* & 0.259 & 0.520 & 0.241 & 0.403 & 0.716  \\ 
                       & \emph{Gaussian Decoder}  & 0.494 & 0.846 & 0.539 & 0.623 & 0.967 \\
                       & \emph{Gaussian Layer}  & \textbf{0.506} & \textbf{0.862} & \textbf{0.548} & \textbf{0.634} & \textbf{0.975}
                       \\ \hline
CAPG-GP                & Goldman et. al* & 0.431 & 0.684 & 0.519 & 0.481 & 0.721  \\
                       & \emph{Gaussian Decoder}  & 0.482 & \textbf{0.782} & 0.573 & 0.542 & \textbf{0.819} \\
                       & \emph{Gaussian Layer}  & \textbf{0.510} & 0.777 & \textbf{0.616}   & \textbf{0.572} & 0.816
                       \\ \hline
\end{tabular}
\caption{Performance of $our$ approach across different general product datasets. * denotes results obtained using the trained model given at \href{https://github.com/eg4000/SKU110K_CVPR19/issues/9}{URL} as is.}
\label{table:others}
\end{table*}

\subsection{Comparision on Other Datasets}
We also compare our trained model on different datasets  \cite{WebMarket}  \cite{GP}  \cite{FGC-OSL}  \cite{holoselecta}, unlike SKU-110k these datasets are not that dense, the number of ground truth bounding box per image are 37, 13, 20 and 34 respectively. We want to clarify that we have not fine tuned our model on any of these datasets and while training there were no augmentations with respect to different scale and size.  \cite{varadarajan2019benchmark} has given a detailed analysis on these datasets with the general object annotations which we use to compare our model accuracy. As shown in Table \ref{table:others}, our Gaussian Decoder and Gaussian Layer Network outperforms the model given by  \cite{sku110k} on WebMarket \cite{WebMarket}, Grocery Products \cite{GP} and CAPG-GP \cite{FGC-OSL} dataset by a large margin, where as we see a drastic performance loss in the Holoselecta \cite{holoselecta} dataset. The performance drop in the Holoselecta dataset is observed because of their varied image dimensions and the object scale variance in the datasets, these mistakes can be seen in Fig. \ref{fig:sample2}. This can be easily solved with multi-scale testing or training but we perform single scale testing on all the datasets for fair comparison.

\section{Conclusion}
\begin{figure}
  \includegraphics[width=\linewidth]{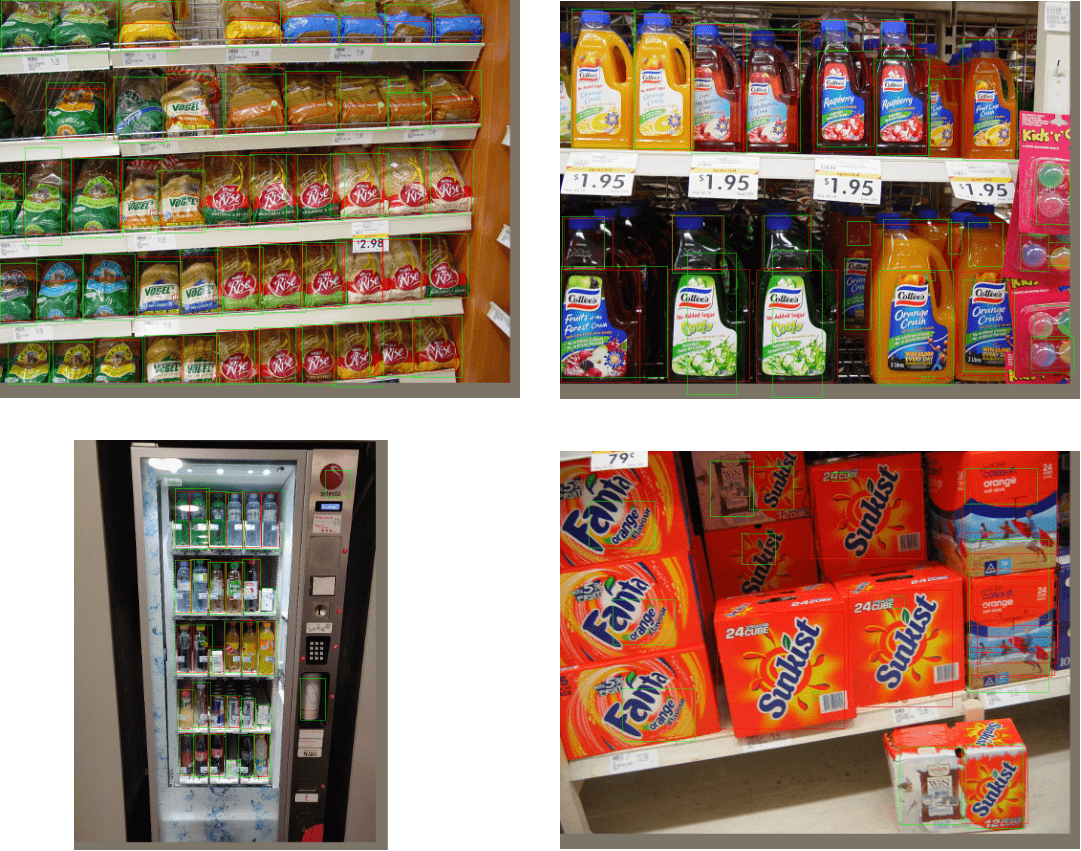}
  \caption{\textbf{Example Prediction on other Datasets by GLN}. (Left column) Examples where the network performance is good. (Right column) Examples where the network performance is poor. These examples show that the trained network doesn't perform well when there are close-up images but has a good performance when the images are taken from a distance.}
  \label{fig:sample2}
\end{figure}

In this work, we proposed an additional multi-task training on the existent RetinaNet architecture. As shown in Fig.  \ref{fig:sample} gaussian layer network does not confuses with the background as much as the simple RetinaNet because of the gaussian map training, the network now is more robust to background objects and can distinguish better between objects placed in close proximity. This gives the significant boost in accuracy in various datasets without any overhead. Our proposed gaussian decoder network shows the affect of multitask training with shared encoder whereas gaussian layer network shows the same with shared encoder and decoder. The improvement in accuracy from gaussian decoder to gaussian layer network also proves our hypothesis of having shared representations for the anchors. We have also shown some results of our trained network on other datasets in Fig. \ref{fig:sample2}

\bibliography{egbib}
\end{document}